\documentclass{article}

\usepackage[margin=1in]{geometry}

\usepackage{helvet}

\usepackage{graphicx}
\usepackage{url}            %
\usepackage{booktabs}       %
\usepackage{amsfonts}       %
\usepackage{nicefrac}       %
\usepackage{amsmath}
\usepackage{amssymb}
\usepackage{float}
\usepackage{lipsum}
\usepackage{booktabs} 
\usepackage{subcaption}
\usepackage[table]{xcolor}
\usepackage{latexsym}
\usepackage{enumitem}
\usepackage{tablefootnote}
\usepackage[round,semicolon]{natbib}
\usepackage{xspace}
\usepackage{textcomp}
\usepackage{makecell}
\usepackage{multirow}
\usepackage{lscape} 
\usepackage{siunitx}
\usepackage{microtype}
\usepackage{url}            %
\usepackage{booktabs}       %
\usepackage{amsfonts}       %
\usepackage{nicefrac}       %
\usepackage{changepage}
\usepackage{xargs}          %
\usepackage{wrapfig,lipsum,booktabs}
\usepackage{enumitem}
\usepackage{longtable}
\usepackage{makecell}
\usepackage{graphicx}
\usepackage{subcaption}
\usepackage[T1]{fontenc}
\usepackage{tgpagella}
\usepackage{xspace}  %
\usepackage{pbox}
\usepackage{fdsymbol}

\usepackage{tikz}

\definecolor{mydarkblue}{rgb}{0,0.08,0.45}

\definecolor{lightblue}{rgb}{0.83, 0.92, 0.95} %
\usepackage[colorlinks,citecolor=mydarkblue,urlcolor=mydarkblue,linkcolor=mydarkblue]{hyperref}

\usepackage[most]{tcolorbox}
\usepackage{bm}
\title{
\vspace{-2em}%
  \vspace{0.1em}%
  \center
  \vskip 0.4in%
  \vskip -\parskip%
 \fontsize{20}{20}{\textbf{Reka Core, Flash, and Edge: A Series of Powerful Multimodal Language Models
}}
  \vskip -\parskip%
  \vskip 0.01in}

\vspace{15em}
 
\author{
\normalsize{}
 \textbf{
 Aitor Ormazabal \hspace{3mm} Che Zheng \hspace{3mm} Cyprien de Masson d’Autume \hspace{3mm} Dani Yogatama}  \vspace{3mm}  \\
 \normalsize{}
 \textbf{Deyu Fu \hspace{3mm} Donovan Ong \hspace{3mm} Eric Chen \hspace{3mm} Eugenie Lamprecht \hspace{3mm} Hai Pham \hspace{3mm} Isaac Ong} \vspace{3mm} \\
 \normalsize{}
 \textbf{Kaloyan Aleksiev \hspace{3mm} Lei Li \hspace{3mm} Matthew Henderson \hspace{3mm} Max Bain \hspace{3mm} Mikel Artetxe} \vspace{3mm}  \\ 
 \normalsize{}
 \textbf{Nishant Relan \hspace{3mm} Piotr Padlewski \hspace{3mm} Qi Liu \hspace{3mm} Ren Chen \hspace{3mm}
 Samuel Phua} \vspace{3mm}  \\
 \normalsize{}
  \textbf{Yazheng Yang \hspace{3mm} Yi Tay \hspace{3mm} Yuqi Wang \hspace{3mm} Zhongkai Zhu \hspace{3mm} Zhihui Xie} \\
 \vspace{3em}\\
 }

\date{}
\begin{document}

\maketitle

\begin{figure}[h]
    \vspace{-4.5em}
    \centering
    \includegraphics[width=0.35\linewidth]{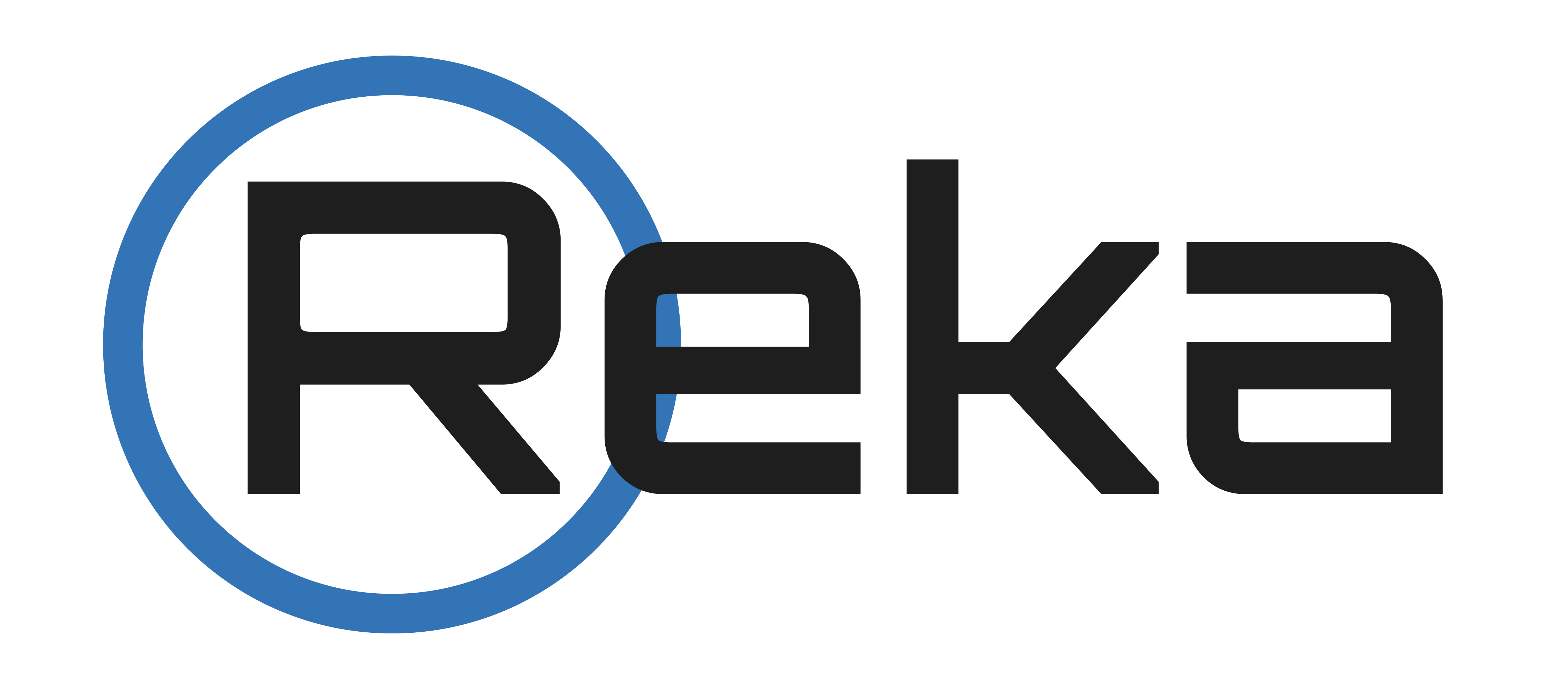}
    \label{fig:enter-label}
\end{figure}
\vspace{1em}
\begin{abstract}
We introduce Reka Core, Flash, and Edge, a series of powerful multimodal language models trained from scratch by Reka.\footnote{Please cite this report as authored by Reka team.} Reka models are able to process and reason with text, images, video, and audio inputs. This technical report discusses details of training some of these models and provides comprehensive evaluation results. We show that \textbf{Reka Edge} and \textbf{Reka Flash} are not only state-of-the-art but also outperform many much larger models, delivering outsized values for their respective compute class. Meanwhile, our most capable and largest model, \textbf{Reka Core}, approaches the best frontier models \citep{openai2023gpt4,geminiteam2023gemini_short,claude3} on both automatic evaluations and blind human evaluations. On image question answering benchmarks (e.g., MMMU, VQAv2), \textbf{Core} performs competitively to GPT4-V. Meanwhile, on multimodal chat, \textbf{Core} ranks as the second most preferred model under a blind third-party human evaluation setup, outperforming other models such as Claude 3 Opus. On text benchmarks, \textbf{Core} not only performs competitively to other frontier models on a set of well-established benchmarks (e.g., MMLU, GSM8K) but also outperforms GPT4-0613 on human evaluation. On video question answering (Perception-Test), \textbf{Core} outperforms Gemini Ultra. Models are shipped in production at \url{chat.reka.ai}. A showcase of \textit{non cherry picked} qualitative examples can also be found at \href{https://showcase.reka.ai}{showcase.reka.ai}.
\vspace{3em}

\end{abstract}
 \normalsize{}

\newpage

\section{Introduction}
This technical report details  comprehensive evaluations of the Reka models (Core, Flash, Edge) on language and vision tasks along with discussions on development, benchmark design, and the training pipeline.

\begin{figure}[H]
    \centering
        \caption{Price per performance (MMLU score) of different LLM APIs. }
    \includegraphics[width=0.8\linewidth]{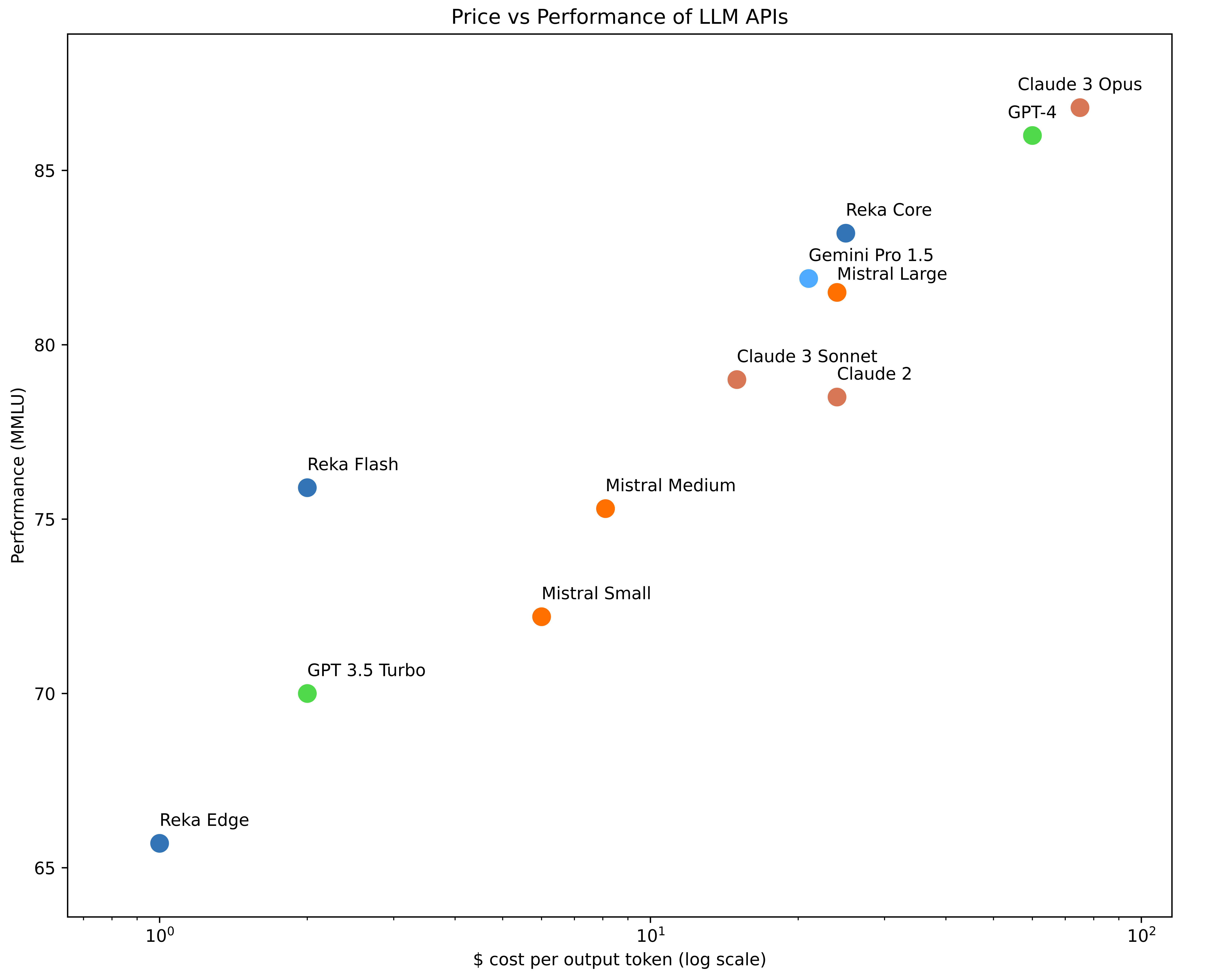}
    \label{fig:priceperf}
\end{figure}

Reka \textbf{Edge} and \textbf{Flash} are dense models with 7B and 21B parameters, respectively. Our evaluation shows that these models are state-of-the-art for their compute class, often surpassing models much larger. Meanwhile, the current version of Reka \textbf{Core} approaches many of the best frontier models~\citep{openai2023gpt4,geminiteam2023gemini_short,google2023palm,claude3}. It excels in both automated base model evaluations and blind third-party human evaluations. Figure \ref{fig:priceperf} compares Reka models against proprietary large language models (LLM) APIs. We plot the price against performance, using MMLU score as an approximate indicator of model quality. All Reka models are positioned either on or beyond the Pareto frontier.

Reka Core approaches the performance levels of GPT-4V \citep{gpt4v} on MMMU \citep{yue2023mmmu}, VQAv2, and third-party multimodal chat evaluation. Meanwhile, Reka Core surpasses all Claude 3 models (Opus, Sonnet, Haiku) \citep{claude3} on multimodal chat human evaluation. On video question answering (Perception-test \citep{pătrăucean2023perception}), both Reka Flash and Core outperform Gemini Ultra \citep{geminiteam2023gemini_short}. On language benchmarks, Reka Core achieves $83.2$ MMLU score and competitive GSM8K, HumanEval, and GPQA scores compared to other frontier models. On text-only chat, blind human evaluation shows that Reka Core outperforms GPT-4 ($0613$) and ranks third on our internal ELO leaderboard (right after GPT-4 Turbo and Claude 3 Opus). 

Meanwhile, our \textbf{Edge} (7B) model surpasses the current state-of-the-art models of this compute class, outperforming both Gemma 7B \citep{gemmateam2024gemma} and Mistral 7B \citep{jiang2023mistral}. Additionally, the \textbf{Flash} (21B) model, aside from outperforming GPT-3.5 Turbo, also outperforms much larger state-of-the-art models such as Grok-1 \citep{grok}, Mistral Medium \citep{touvron2023llama} and Gemini Pro 1.0 \citep{geminiteam2023gemini_short}. On multimodal evaluations, \textbf{Flash} outperforms both Claude 3 Opus and Sonnet \citep{claude3} on multimodal chat and matches the Sonnet model on MMMU \citep{yue2023mmmu}. All in all, the \textbf{Edge} \& \textbf{Flash} models are extremely powerful models on a compute-class basis. 

In addition to comprehensive evaluations and benchmark evaluations on both language and vision (video + image) tasks, this report also shares some interesting technical details and behind-the-scenes of training large multimodal models as a startup. Areas discussed include infrastructure, data pipeline, compute, annotation pipelines, and more. Finally, artifacts of our models (playground/chat, developer platform) can be found in the following resource table (Table \ref{tab:resouce_table}).

\begin{table}[H]
\caption{Resource tree of Reka artifacts.}
    \centering
    \begin{tabular}{l|c}
    \toprule
    What & Where? \\
    \midrule
        Playground (chat app) & \href{https://chat.reka.ai}{chat.reka.ai} \\
       Qualitative Examples (static, non-cherry picked)  & \href{https://showcase.reka.ai}{showcase.reka.ai} \\
       API platform (sign up, manage credits) & \href{https://platform.reka.ai/}{platform.reka.ai} \\ 
       Discord (questions) & \href{https://discord.com/channels/1191356083130880041/1191356084682752022}{discord} \\
       Homepage & \href{https://reka.ai}{reka.ai} \\
       \bottomrule
    \end{tabular}
    
    \label{tab:resouce_table}
\end{table}
\vspace{-2em}

\section{Model}
This section briefly describes the technical details behind these models.
\vspace{-1em}
\subsection{Training Data}
The training data comprises a mixture of publicly available and proprietary/licensed datasets with a dataset knowledge cutoff of November 2023. The dataset ingested by our model comprises of text, images, videos, and audio clips. Reka Flash and Reka Edge were trained on approximately $5$ trillion and $4.5$ trillion extensively deduplicated and filtered language tokens, respectively. While the classification of corpora is not strictly defined to one class or category, approximately $25\%$ of our pretraining data is code related, and ${30}\%$ are STEM related. Approximately $25\%$ of the data is web crawl. About $10\%$ of our data has some relation to math. Overall mixture rates generally follow a principle of prioritizing unique tokens but are hand-adjusted using signal from a limited number of small scale ablations. 
\begin{table}[H]
\centering
    \caption{Statistics of Reka suite of multimodal language models. \textbf{Note: Reka Core has not finished training and is still improving.}}
    \begin{tabular}{lccccc}
    \toprule
Model &  Model Size & Text tokens & Context & Long-context & Knowldge Cutoff \\
\midrule 
Edge & $7$B dense & $4.5$T & $8$K & $64$K & Nov $2023$  \\
Flash & $21$B dense & $5$T & $8$K & $128$K & Nov $2023$\\ 
Core & - & - & $8$K & $128$K & Nov $2023$ \\
\bottomrule
\end{tabular}
\end{table}

\textbf{Multilingual Data}: 
Approximately $15\%$ of our pretraining data is explicitly (and deliberately) multilingual, comprising $32$ diverse languages tier-weighted (roughly by frequency in the wild). Beyond these explicitly up-weighted languages, we also train on the entire multilingual Wikipedia comprising of 110 languages so we expect a baseline performance for most languages. It is worth noting that these tiers reflect pretraining capability and not necessarily downstream post-training induced capabilities of the final model. To be concrete, these are meaningful to estimate the potential of a particular language, given suitable supervised fine tuning data.
 Languages included during pretraining are shown below.
\begin{table}[H]
\small
\centering
    \caption{Tiered languages in pretraining mixture.}
    \begin{tabular}{lp{10cm}}
    \toprule
    Pretraining Tier & Languages \\ 
    \midrule
P1 languages & German, Chinese, Japanese, French, Korean, Spanish, Italian, Arabic, Hindi \\ 
P2  languages & Indonesian, Vietnamese, Thai, Czech, Dutch, Finnish, Bulgarian, Portuguese, Tamil, Persian, Greek, Russian \\
Additional languages &  Turkish, Telugu, Burmese, Swahili, Urdu, Estonian, Malay, Basque, Swedish, Norwegian\\
\bottomrule
\end{tabular}
\end{table}

\textbf{Multimodal Data}:
The multimodal training data comprises large collections of images, videos, documents, and webpages. The chosen data mixture is carefully optimized for quality, diversity, and scale.

\subsection{Architecture \& Modeling}

\begin{figure}[!hb]
\caption{\textbf{Architectural overview for Reka Core, Flash \& Edge models:} a modular encoder-decoder transformer supporting multimodal input (image, text, video \& audio). The text output can invoke function calls, such as web search and code execution, then return the results.}
    \centering
 \includegraphics[width=0.7\linewidth]{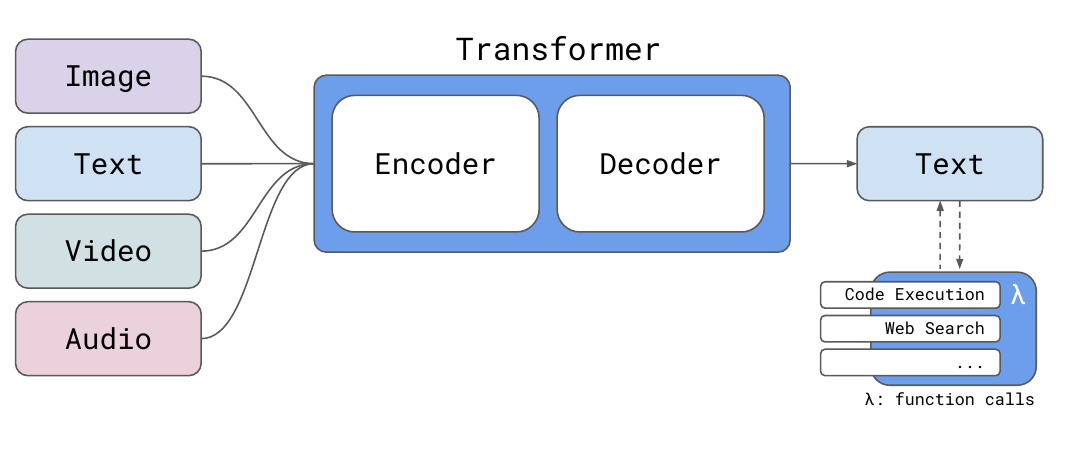}
    
    \label{fig:reka-arch}
\end{figure}

This section introduces training details, model architecture, and context length details. 
\paragraph{Architecture \& Training.} Our overall architecture (Figure~\ref{fig:reka-arch}) is a modular encoder-decoder architecture supporting text, image, video, and audio inputs. For now, our model only supports text outputs. The backbone Transformer model is based on the 'Noam' architecture, i.e., it uses SwiGLU \citep{shazeer2020glu}, Grouped Query Attention \citep{ainslie2023gqa,shazeer2019fast}, Rotary positional embeddings \citep{rope-paper} and RMSNorm \citep{zhang2019root}. Architecturally, this is similar to the PaLM architecture \citep{chowdhery2022palm} but without parallel layers. Reka Flash and Edge uses a sentencepiece vocab of 100K based on \textit{tiktoken} (e.g., GPT-4 tokenizer). We add sentinel tokens for masking spans, i.e., \textsc{<extra\_id\_0>} and other special use cases such as tool-use that are beyond the scope of this technical report. Pretraining uses a curriculum that goes through multiple stages with different mixture distributions, context lengths, and objectives. The current version of this model is a dense model. Models are trained with \textsc{bfloat16}. 
\paragraph{Context Length.} Our standard models have a context length of 8K for our regular models. Reka Flash and Reka Core have $128$K for long context models for retrieval and long document tasks. All our models pass needle-in-the-haystack (passkey retrieval) for the context they support. Based on these tests, our 128K models seem to extrapolate to $256$K context length (but not beyond). For long context training, in addition to instruction tuning data we collect, we synthetically create supervised fine tuning data using our own suite of models by conditioning on long documents found in pretraining corpus using a technique we call \textit{reverse instruction tuning from long documents}.

\subsection{Compute \& Infrastructure} 
\begin{figure}[H]
    \centering
        \caption{Training loss for Reka Core.}
    \includegraphics[width=0.9\linewidth]{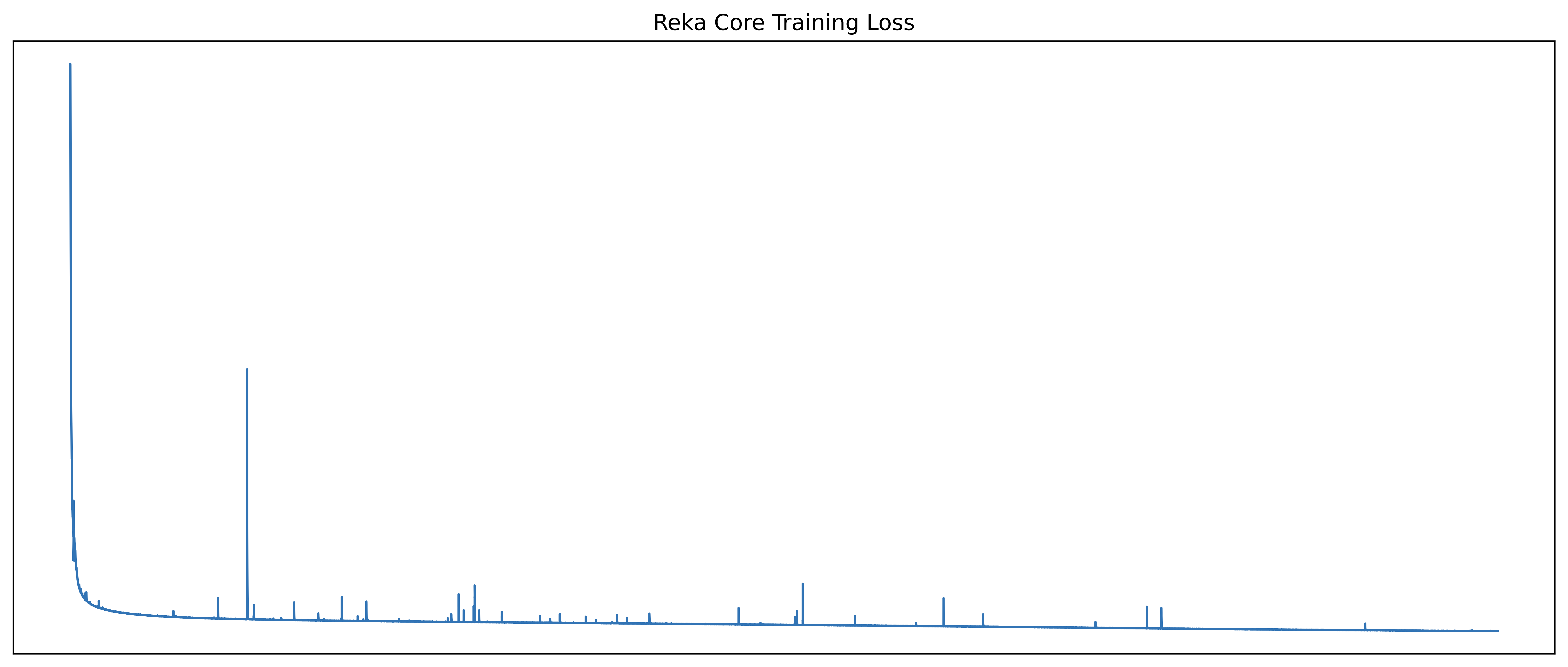}
    \label{fig:coreloss}
\end{figure}

Our family of Reka models was trained predominantly on Nvidia H100s using Pytorch \citep{pytorch}. Our setup comprises of clusters from a mixture of vendors with our peak compute being approximately $2.5$K H$100$s and $2.5$K A$100$s. Our peak number of clusters is $6$. About more than $90\%$ of our compute came online in mid-December 2023. Reka Flash and Edge were trained on several hundreds of H100s across a period of several weeks. Our pretraining process was relatively smooth with very few loss spikes despite very aggressive learning rates\footnote{Models trained at the edge of stability turn out stronger. See \url{https://x.com/m__dehghani/status/1686056450081337344}.} even for much larger models. Figure \ref{fig:coreloss} shows the training loss for Reka Core. To improve the I/O of our clusters, especially for scalable training with multimodal inputs, we used the Ceph filesystem for distributed and scalable data storage across nodes which improved I/O substantially but came with maintenance overheads.

\subsubsection{Hardware lottery and node stability.} 
Generally, we find great unreliability when it comes to GPU nodes which often fail due to hardware errors or connection issues. Moreover, reliability among providers is generally of high variance. For more details, refer to~\citet{tayblog}. To expand upon \citet{tayblog}, we report the average number of node failures across four anonymized providers, as shown in Table \ref{tab:failure_nodes}. Since the likelihood of node failures is influenced by the number of nodes concurrently used for training, we report estimated failure rates for different configurations. 

\paragraph{Chaotic and Stable phases} Aside from variances across clusters and providers, providers could also have high variance across time periods. For example, many compute providers have clusters that behave very differently in the first few weeks of handover or whenever the cluster undergoes a big change. Hence, we also compare the node failure rates during both the early phase and stable phase. More often than not, aside from early phase of handing over a cluster, provisioning new nodes can also introduce a new chaotic era that can last a few days or weeks. %
In general, we determined that a key factor influencing the difference between the early and stabilized phases is whether the cluster was actively used for distributed training by previous customers.
\begin{table}[H]
    \centering
     \caption{Average number of node failures (on a weekly basis) across four anonymized compute providers. Since node failures depend on the number of nodes used concurrently, we report estimated failure rates for different configurations. Many compute providers have clusters that behave very differently in the first few weeks of handover. Hence, we also report the difference in node failure rate in both the early phase and stable phase. Chips refer to either H100s or A100s.}
    \begin{tabular}{lp{2cm}p{3cm}}
    \toprule
   Provider      &  Number of chips used &  Number of node failures (per week) \\
   \midrule
        Provider A  & $2000$ chips & $3$ \\ 
        Provider A (early phase)  & $2000$ chips & $20+$ \\ 
        \midrule
        Provider B & $300$ chips & $0.2$ \\ 
        Provider B (early phase) & $300$ chips & $0.2$ \\ 
        \midrule
        Provider C (stable phase) & $300$ chips & $3$ \\ 
    Provider C (stable phase) & $100$ chips & $3$ \\ 
        Provider C (early phase) & $100$ chips & $30+$ \\ 
        \midrule
        Provider D & $300$ chips & $2$ \\ 
\bottomrule
    \end{tabular}
   
    \label{tab:failure_nodes}
\end{table}

\paragraph{Inference and serving.} We built a custom inference stack for text and multi-modality running on a combination of A10s and A100s. We use Kubernetes as the underlying orchestration engine and manage several large clusters across different regions.

\subsection{Post-Training}
This section describes the post-training process which involves aligning, instruction tuning the model.
\paragraph{SFT and RLHF.} After pretraining, our models are then instruction tuned \citep{wei2021finetuned,ouyang2022training,chung2022flan} for multiple epochs using strong regularization. As for SFT data, we train on a mixture of datasets that include our proprietary and publicly available data. After SFT, models are then aligned with RLHF, specifically PPO \citep{ppo}, using the same family of Reka models as the reward model. Our models go through a couple of rounds of RLHF in total. Moreover, our post-training process considers tool-use, function calling and web search, which is out of scope for this technical report.

\paragraph{Annotation Pipelines for Data Collection and Human Evaluation.} We collect data using external data collection companies and provide them with a user interface for annotating both text-only and multimodal data. We create an annotation UI for both collecting data and/or sending examples to human raters for blind human evaluation. This software also supports annotating for individual pointwise quality and also side-by-side (pairwise) evaluations. Our annotation software supports images, videos, and text-only prompts and responses. It also supports the annotation of multi-turn dialogues.

\section{Evaluation}
This section discusses the results of extensive evaluations of Reka models.

\subsection{Base Model Evaluation}
We conduct a series of language-only and multimodal (image, video input) evaluations. 
\paragraph{Language Model Evaluation.} We compare our models on four language model evaluations: 1) MMLU (general language understanding and question answering) \citep{hendrycks2021measuring}, 2) GSM8K (reasoning and arithmetic) \citep{cobbe2021training}, HumanEval (code generation) \citep{chen2021codex} and GPQA (graduate-level question answering) \citep{rein2023gpqa}. All numbers from baselines are reported numbers in other works. MMLU is evaluated with 5-shot direct prompting for all models. For GSM8K, most models use 8-shot chain-of-thought \citep{wei2022chain} and majority voting (maj@8). For HumanEval, this is evaluated in 0-shot setup. All results from other models are reported from other works. 

\paragraph{Multimodal (Image/Video) Evaluation.} We compare our models using visual question answering datasets, i.e., MMMU \citep{yue2023mmmu}, VQAv2~\citep{balanced_vqa_v2}, and Perception-Test \citep{pătrăucean2023perception} for video question answering. For Reka models, all results are 0-shot. 

\newcolumntype{H}{>{\setbox0=\hbox\bgroup}c<{\egroup}@{}}
\begin{table}[ht]
    \centering
    \caption{Comparisons of our Reka Flash and Reka Core against other frontier models. Dashes ($-$) refer to either model not supporting modality or unavailable benchmark scores.}
    \renewcommand{\arraystretch}{1.2}
    \begin{tabular}{p{2.4cm}|p{1.6cm}p{1.65cm}|Hp{1.6cm}p{1.5cm}p{1.5cm}p{1.5cm}p{1.5cm}p{1.5cm}}
    \toprule
 Model / Eval & Reka Core v$0.5$ &  Reka Flash v$1.5$  & Edge & GPT-4 & Claude 3 Opus & Claude 3 Sonnet &  Gemini Ultra & Gemini Pro $1.5$\\ 
 \midrule 
 MMLU \newline
 (\textit{Knowledge})& $83.2$ & $75.9$ & $65.4$ & $86.4$ & $86.8$ & $79.0$ & $83.7$ & $81.9$ \\
 GSM8K \newline (\textit{Reasoning}) &  $92.2$ & $85.8$  & $72.1$ & $92.0$ & $95.0$ & $92.3$ & $94.4$ & $91.7$\\
 HumanEval \newline (\textit{Coding})  & $76.8$ &  $72.0$ & $55.5$ & $76.5$ & $84.9$& $73.0$ & $74.4$ & $71.9$\\
 GPQA (\textit{main}) \newline(\textit{Hard QA}) & $38.2$ & $34.0$ & & $38.1$ & $50.2$ & $39.1$&   $35.7$ & $41.5$\\ 
 \midrule 
 MMMU \newline (\textit{Image QA}) & $56.3$ & $53.3$ & & $56.8$ & $59.1$ & $53.1$ & $59.4$ & $58.5$  \\
 VQAv2 \newline (\textit{Image QA}) & $78.1$ & $78.4$ & & $77.2$ & $-$ & $-$ & $77.8$ & $73.2$ \\
 Perception-test (\textit{Video QA}) & $59.3$ & $56.4$  & & $-$  & $-$ & $-$& $54.7$ & $51.1$\footnotemark \\ 
 \bottomrule
    \end{tabular}
    \label{tab:text_benchmarks}
\end{table}

\footnotetext{We report Pro 1.0 performance here since Pro 1.5 did not report perception-test.}

\paragraph{Results.} Table \ref{tab:text_benchmarks} reports comparisons of Reka Core against other frontier-class models. Overall, Reka Core performs competitively with other frontier-class models. On most metrics (with the exception of MMLU), it is comparable to GPT-4\footnote{At least an older version, with the results mostly reported from the recent Claude 3 release. HumanEval looks too low for the Claude 3 release so we referenced the HumanEval leaderboard for this number.}. In terms of overall performance and with respect to the Claude 3 series, it falls somewhere in between Opus and Sonnet. When compared to Gemini models, Reka Core has mixed outcomes, i.e., winning some and losing some. Reka Core outperforms Gemini Pro 1.5 on several benchmarks (MMLU, GSM8K, HumanEval) but is outperformed on GPQA and MMMU. Notably, Reka Core and Flash outperform Gemini Ultra (and Pro 1.5) on video question answering. Reka Core is still improving so we expect better results in the near future.

\subsection{Chat Model Evaluation}
We conduct a blind evaluation with human raters from a third party data provider company. We consider two setups: 1) multimodal chat, where the user asks a question about an image, and 2) text-only chat. We next detail our evaluation protocol and present results for each setting.

\subsubsection{Evaluation Setup}

For each annotation instance, human raters are given a prompt along with a maximum of 4 generations from different models, and asked to rate the answers according to the guidelines provided. Given that the number of models in our evaluation is higher than 4, we collect multiple such annotations for each prompt, each with a different subset of models. The pairing of models is decided randomly for each prompt, with all combinations being equally likely. We compute ELO scores following \citet{askell2021general}, where we only consider pairwise comparisons where annotators express a preference stronger than the weakest available.

We design our evaluation dataset to cover a diverse set of prompts. The following table details the composition of our text-only evaluation set, which comprises 1K+ prompts:

\begin{table}[H]
\small
\centering
    \caption{Taxonomy of prompts in our text-only human evaluation dataset. The dataset is balanced across subcategories.}
    \begin{tabular}{ll}
    \toprule
\textbf{Category} & \textbf{Subcategory} \\
\midrule

\multirow{5}{*}{Knowledge-intensive}
& Humanities and social sciences \\
& Natural sciences \\
& Engineering and technology \\
& Entertainment \\
& Other \\

\midrule

\multirow{5}{*}{Creative writing}
& Role playing \\
& Brainstorming \\
& Poetry \\
& Literary prose \\
& Non-literary prose \\

\midrule

\multirow{7}{*}{Input-based}
& Data processing \\
& Reading comprehension \\
& Classification \\
& Extraction \\
& Summarization \\
& Rewriting \\
& Translation \\

\midrule

\multirow{3}{*}{Reasoning}
& Maths \\
& Commonsense and logical reasoning \\
& Instruction following \\

\midrule

Coding & N/A \\

\bottomrule
\end{tabular}
\end{table}

Similarly, the following table reports the categories covered by our multimodal evaluation set:

\begin{table}[H]
\centering
    \caption{Distribution of prompts in our multimodal human evaluation dataset.}
    \small
    \begin{tabular}{lr}
    \toprule
\textbf{Category} & \textbf{Ratio} \\
\midrule

Basic image description & 23.0\% \\
Advanced image description & 20.5\% \\
Coding capability with vision & 7.7\% \\
Multilingual multimodal understanding & 7.9\% \\
Multimodal knowledge and commonsense & 7.7\% \\
Scene and document reasoning & 13.0\% \\
Visual referring prompting & 5.1\% \\
Creative tasks & 2.6\% \\
Other & 12.5\% \\

\bottomrule
\end{tabular}
\end{table}

\subsubsection{Multimodal Chat Evaluation}

We next report the results of our multimodal chat evaluation in comparsion with GPT4-V, Claude 3, Gemini Pro, IDEFICS 80B, Adept Fuyu 8B, and the strongest Llava 1.6B model:

\begin{table}[H]
\centering
    \caption{ELO scores of all models on our multimodal human evaluation.}
    \label{multimodalhumanevaltable}
    \begin{tabular}{lrr}
    \toprule
Model &  ELO  & Win rate \\
\midrule 
GPT-4V & 1201 & 79.4 \\
\textbf{Reka Core} & 1130 & 72.2 \\
\textbf{Reka Flash} & 1082 & 66.8 \\
Claude 3 Opus & 1073 & 66.2 \\
Claude 3 Sonnet & 1069 & 64.1 \\
Llava 1.6 34B & 1022 & 55.9 \\
Gemini Pro & 1011 & 54.2 \\
\textbf{Reka Edge} & 986 & 50.5 \\
IDEFICS 80B & 732 & 18.8 \\
Adept Fuyu 8B & 550 & 6.4 \\
\bottomrule
\end{tabular}
\end{table}

We find that Reka Core outperforms all models except GPT4-V by a substantial margin. Reka Flash ranks next, performing marginally better than Claude 3 Opus. Reka Edge outperforms IDEFICS 80B and Adept Fuyu 8B by a large margin, approaching the performance of Gemini Pro and the largest Llava 1.6 model.

\subsubsection{Text-only Chat Evaluation}

We compare our models against different versions of GPT, Claude 3, Llama 2 Chat, and Gemini Pro (API version), and report our results next:

\begin{table}[H]
\centering
    \caption{ELO scores of all models on our text-only human evaluation.}
    \begin{tabular}{lrr}
    \toprule
Model &  ELO  & Win rate \\
\midrule 
GPT-4 Turbo (1106-preview) & 1227 & 78.6 \\
Claude 3 Opus & 1185 & 73.6 \\
\textbf{Reka Core} & 1091 & 60.6 \\
Claude 3 Sonnet & 1074 & 59.0 \\
GPT-4 (0613) & 1062 & 57.0 \\
\textbf{Reka Flash} & 1020 & 49.1 \\
GPT-3.5 Turbo (0613) & 1012 & 48.9 \\
Llama 2 Chat 70B & 984 & 43.0 \\
Gemini Pro & 950 & 38.3 \\
\textbf{Reka Edge} & 903 & 31.5 \\
Llama 2 Chat 7B & 850 & 24.3 \\
\bottomrule
\end{tabular}
\end{table}

We find that Reka Core ranks competitively on our ELO leaderboard, outperforming Claude 3 Sonnet and GPT-4, and it is only surpassed by GPT-4 Turbo and Claude 3 Opus. Reka Flash obtains strong results for its size, beating GPT-3.5 Turbo, Gemini Pro and the much larger Llama 2 Chat 70B.

\subsubsection{Model development and automatic evaluation using Reka Core}
We leverage the frontier-class capabilities of Reka Core for model selection and development and show an example of how we use it for multimodal chat. We ask Reka Core to simulate human judgement by rating a response with respect to a  prompt and a reference answer. In short, $f(\text{prompt}, \text{model\_output}, \text{reference\_answer}) \in [1,100]$. We find that Reka Core rankings across models correlate to human judgement despite the gap between pointwise and pairwise (arena style) evaluations. Our general workflow is that we perform lightweight and simple \textit{pointwise} evaluations for continuous sanity checks before sending our models for third party blind human evaluations. 

\begin{figure}[H]
\caption{Results using Reka Core as an evaluator. Reka Core evaluator scores align almost perfectly with the final ELO scores we obtain from human raters.}
    \centering
    \includegraphics[width=0.75\linewidth]{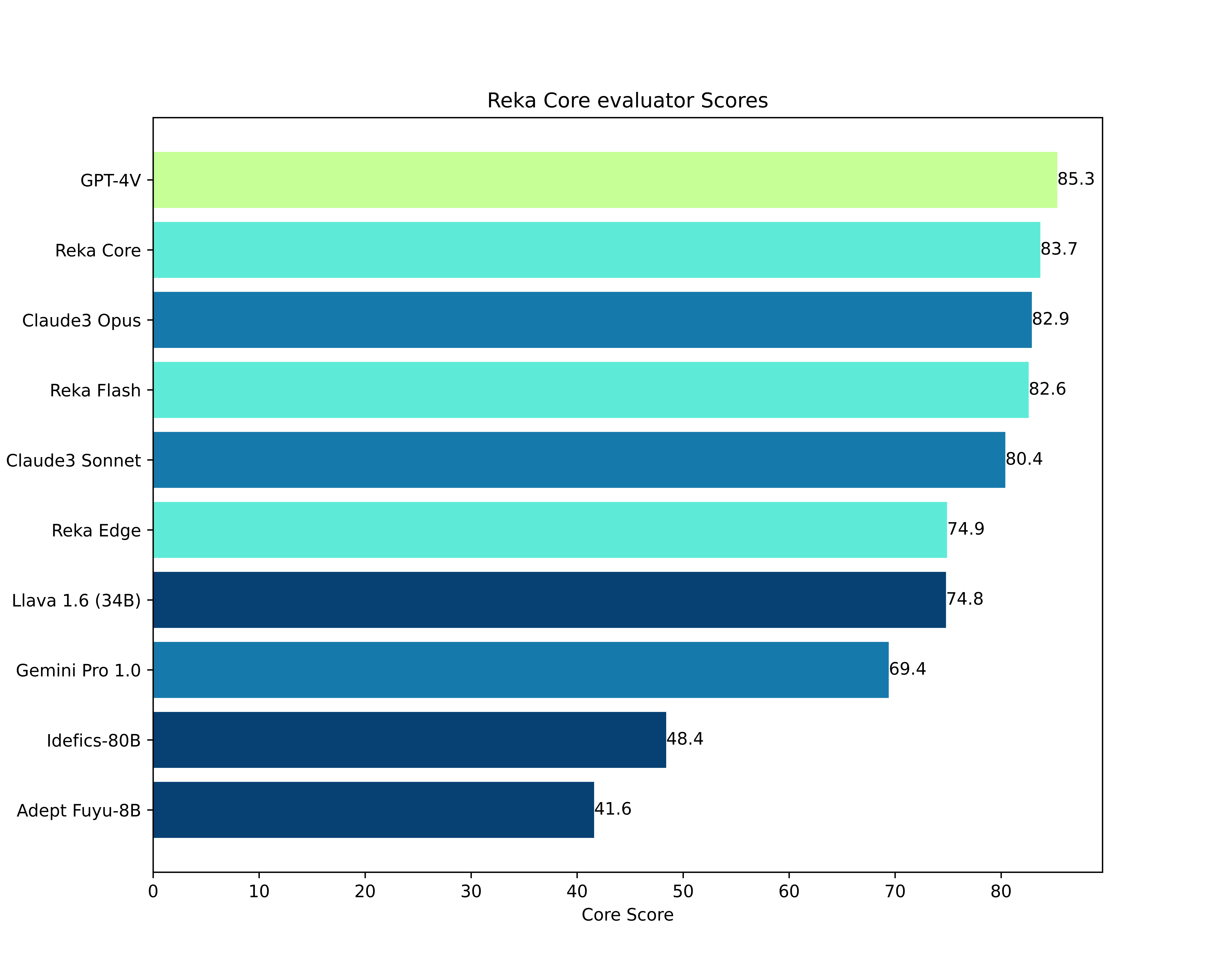}
     
    \label{fig:coreeval}
\end{figure}

Figure \ref{fig:coreeval} reports the Reka Core scores we obtain right before producing Table \ref{multimodalhumanevaltable}. Despite Reka Core evaluations being pointwise, we find that it is able to accurately approximate the final rankings. Here, the only key difference is that Reka Flash and Claude Opus have flipped rankings. In practice, these models may be very similar in performance that it could go either way. In Table, \ref{multimodalhumanevaltable}, we also note that Reka Flash and Claude Opus have very similar win rates and ELO scores, which is well reflected by their Reka Core scores being very close as well. Overall, we find that Reka Core is quite a good approximator of final human evaluation outcomes. 

\subsection{Cross-lingual Evaluations}
We conduct experiments on a suite of general multilingual benchmarks such as multilingual commonsense (XStoryCloze \citep{lin-etal-2022-shot}), causal reasoning (XCOPA \citep{ponti-etal-2020-xcopa}), question answering (Belebele \citep{bandarkar2023belebele}, XQuAD \citep{Artetxe:etal:2019}, TydiQA \citep{clark2020tydi}). For all datasets, we report the mean across all languages. We compare our models with Llama 2 70B \citep{touvron2023llama}, GPT-3.5 and GPT-4. All evaluations are zero-shot generative except XStoryCloze which uses log-likehood evaluation.

\begin{table}[H]
    \centering
        \caption{Statistics of multilingual datasets.}
    \label{tab:multilingual_stats}
    \begin{tabular}{l|p{8cm}p{1cm}}
    \toprule
        Eval & Languages  & Num Langs\\
        \midrule
         XStoryCloze &  hi, te, en, zh, ru, my, sw, es, id, eu, ar & 11\\ 
         XCOPA & sw, th, tr, et, vi, qu, id, zh, it, ta, ht & 11 \\ 
         XQuAD & ar, de, el, en, es, hi, ro, ru, th,tr, vi, zh & 12\\
         XWinograd & fr, en, jp, pt, zh, ru & 6 \\
         TydiQA & ar, bg, en, fi, id, jp, ko, ru, sw, te, th& 11\\
         Belebele & \textit{too many} & 150 \\ 
         \bottomrule 
    \end{tabular}

\end{table}

\begin{table}[H]
    \centering
    \caption{Comparisons of our models on multilingual tasks against GPT-3.5 and GPT-4. All tasks are zero-shot.}
    \begin{tabular}{p{4cm}p{1.2cm}p{1.4cm}p{1.4cm}|p{1.4cm}Hp{1.4cm}p{1.4cm}p{1.4cm}p{1.4cm}}
    \toprule
  Eval / Model & Metric & Reka Core v$0.5$ & Reka Flash v$1.5$ & Llama-2 70B  & Mixtral  & GPT-3.5 & GPT-4\\
    \midrule
  XStoryCloze & acc & \textbf{$72.0$} & $70.1$ & $63.2$ & $62.6$ & N/A & N/A \\ 
  XCOPA & acc & $88.3$ & $68.0$ & $50.6$ & & $72.2$ &$86.3$ \\ 
  XQuAD & EM & $65.7$ & $61.4$ & $25.5$ & $32.1$ & $34.6$  & $44.2$ \\ 
  XWinograd & acc &  $86.8$  & $84.0 $& $65.3$ & $78.9$  & $72.2$ & $91.5$\\ 
  TydiQA (w context)& EM &$60.4$ & $64.8$ & $34.9$ & $28.3$ & $53.1$ & $58.9$\\
  TydiQA (w/o context) &  EM & $17.4$ & $15.7$ & $3.9$ & $8.7$ & $13.5$ &$21.1$ \\
  Belebele (all langs) & acc & $63.4$ & $57.3$ &$48.0$ & $ 39.8$ &  $51.1$ & N/A\\ 
         \bottomrule
    \end{tabular}
    \label{tab:multilingual_benchmarks}
\end{table}

\begin{figure}[H]
    \caption{Comparison of Reka Core vs GPT-4. Breakdown of languages on 0-shot TydiQA (with context).}
    \centering
    \includegraphics[width=1.0\linewidth]{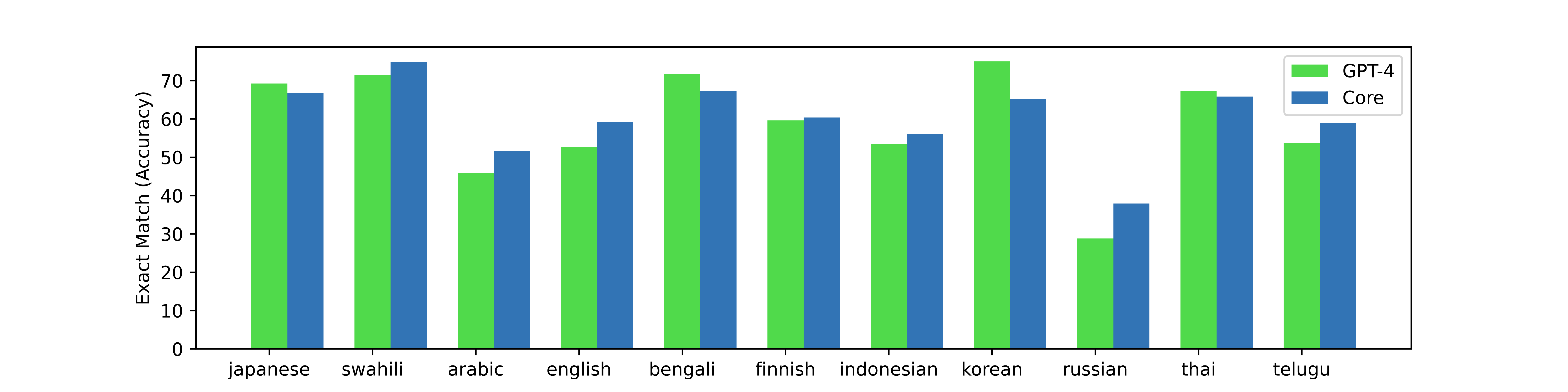}
    \label{fig:tydiqa}
\end{figure}

Table \ref{tab:multilingual_benchmarks} reports our evals\footnote{We do not run evals for GPT models on XStoryCloze because we use logprobs. As for Belebele, we hit our credit threshold just evaluating on this large evaluation dataset so we stopped.} on multilingual benchmarks. Generally we find that Reka Core outperforms all baselines reliably on most tasks (except GPT-4 where it is mixed). Specifically, Reka Core outperforms GPT-4 on XCOPA, XQuAD, TydiQA but is outperformed on XWinograd and TydiQA (w/o context). Meanwhile, Core outperforms Flash on all benchmarks. Both Flash and Core outperforms Llama-2 70B and GPT-3.5. Finally, Figure \ref{fig:tydiqa} shows the language breakdowns of Core vs GPT-4. 

\subsection{Long Context Question Answering}
We conduct a series of evaluations on long context question answering. We use internal benchmarks in two domains: (1) movie plots and (2) ToS (terms-of-service) contract with contexts in the ballpark of $100$K tokens. Both datasets are question answering tasks where the task is to answer questions given a long document. We compare with Claude 3 (Haiku and Sonnet). 

\begin{table}[H]
    \centering
      \caption{Long context question answering evaluation results.}
    \label{tab:long_qa}
    \begin{tabular}{c|cccc}
    \toprule
    Model     &   Reka Core & Reka Flash & Claude 3 Haiku & Claude 3 Sonnet  \\
    \midrule
        Movie Plots & $83.6$ & $79.7$ & $76.6$ & $82.2$ \\
        ToS Contract & $87.5$ & $90.0$ & $85.0$ & $90.0$\\
        \bottomrule
    \end{tabular}
\end{table}
Table \ref{tab:long_qa} reports results on long context question answering using internal evaluation datasets. Overall we show that Flash and Core are both competitive to the latest Claude 3 models.

\subsection{Medical Reasoning}
We compare our Reka models against state-of-the-art domain-specific medical models such as Meditron \citep{chen2023meditron70b} and Med-PaLM-2 \citep{singhal2023expertlevel}. We also compare with GPT-4 reported from \citep{singhal2023expertlevel}. We compare on three benchmarks: MedMCQA, PubMedQA and MMLU (Medical). MMLU medical is a macro-average over \textit{clinical knowledge}, \textit{medical genetics}, \textit{anatomy}, \textit{professional medicine}, \textit{college biology} and \textit{college medicine}. 

\begin{table}[H]
    \centering
        \caption{Results on medical reasoning tasks compared to domain specialized models and frontier models.}
    \label{tab:medical_reasoning}
    \begin{tabular}{p{3.5cm}|p{1cm}p{1cm}p{1cm}|p{1cm}p{1cm}|p{2.4cm}p{2.4cm}}
    \toprule
         &  \multicolumn{3}{c}{Reka} &  \multicolumn{2}{c}{Meditron}   &  & \\
      
         Benchmark / Model  & Edge (7B) & Flash (21B) & Core & 7B & 70B & Med-PaLM-2 & GPT-4 \\ 
            \midrule
            MedMCQA &  $52.6$ & $71.3$ & $80.6$ & $28.7$ & $52.0$ & $71.3$ & $72.4$\\ 
            PubMedQA & $71.6$ & $69.0$ & $74.6$ & $69.3$ & $79.8$ & $79.2$ & $80.4$\\ 
            MMLU (Medical) & $65.7$ & $79.5$ & $88.3$ & $54.2$&$72.7$ &$87.8$ & $90.3$  \\
            \midrule
            Avg & $63.3$ & $73.2$ & $81.3$ & $50.7$ & $68.2$ & $79.4$  & $81.0$\\ 
            \bottomrule
    \end{tabular}

\end{table}

Table \ref{tab:medical_reasoning} reports results on medical tasks. Meditron and Med-PaLM-2 are specialized models for medicine. Our results show that Reka Core is competitive with some of the best frontier models and specialized models in medicine. Firstly, Reka Flash and Core outperforms the Meditron series. Secondly, Reka Core outperforms both Med-PaLM-2 and GPT-4 on MedMCQA. However, it is outperformed on PubMedQA. Finally, on MMLU (Medical), Reka Core outperforms Med-PaLM-2 and is slightly behind GPT-4. Overall, on average, Reka Core outperforms Med-PaLM-2 and is approximately similar to GPT-4 on medical tasks.

\subsection{Detailed comparisons of Edge and Flash}
We report detailed results of Reka Edge and Flash against other models of similar compute class. Notably, both Edge and Flash have been improved quite substantially since the initial release in Feb. Hence, numbers have been upgraded since their first appearances. 
\subsubsection{Reka Edge results}
We report results of Reka Edge against other 7B models such as Llama 2 \citep{touvron2023llama}, Mistral \citep{jiang2023mistral} and Gemma \citep{gemmateam2024gemma}. 
\begin{table}[H]
    \centering
       \caption{Results comparing Reka Edge with other leading 7B models in the industry. Most benchmarks are reported from other works with the exception of those denoted with $\dagger$. For multilingual benchmarks, we run them ourselves.}
    \label{tab:7bresults}
    \begin{tabular}{llcccc}
    \toprule
        Benchmark & metric & Llama 2 7B & Mistral 7B & Gemma 7B & Reka Edge \\
        \midrule
        MMLU & $5$-shot & $45.3$ & $62.5$ & $64.3$ & \textbf{$65.7$}\\ 
         GSM8K &maj@$1$ & $14.6$ & $35.4$& $46.4$ & $66.2$ \\ 
        MATH & $4$-shot & $2.5$& $12.7$& $24.3$ & $23.2$\\
        HumanEval & 0-shot (pass@1) & $12.8$ & $26.2$ & $32.3$ &  $54.3$ \\
        \midrule
        XQuAD$^\dagger$ & 0-shot & $16.6$ & $29.7$ & $21.7$ & $54.2$\\ 
        TydiQA$^\dagger$ & 0-shot & $16.4$ &  $31.7$& $35.8$& $61.5$\\
        TydiQA$^\dagger$ (w/o context) & 0-shot &$2.8$ & $5.0$ & $4.7$ & $6.9$\\ 
        Belebele$^\dagger$ & 0-shot & $27.7$ & $32.8$ & $26.8$ & $37.1$ \\
       
        \bottomrule 
    \end{tabular}
 
\end{table}
Table \ref{tab:7bresults} reports results of Reka Edge against other 7B models (Gemma, Mistral, Llama). We observe that Reka Edge has an edge against all other models (no pun intended). It outperforms Mistral 7B and Llama 7B on all \textbf{8} benchmarks. As for Gemma, it outperforms Gemma for all benchmarks except MATH. Overall, Reka Edge is a super strong model at 7B scale.

\subsubsection{Reka Flash results}
Given that there are not many good models around the same compute class as Reka Flash, we compare Reka Flash with models that are much larger. Specifically, Llama 2 70B \citep{touvron2023llama}, Gemini Pro 1.0 \citep{geminiteam2023gemini_short}, Mistral Medium \citep{touvron2023llama} and Grok-1 \citep{grok}.  
\begin{table}[H]
    \centering
       \caption{Results comparing Reka Flash with other much larger models.}
    \label{tab:flash_results}
    \begin{tabular}{p{2.6cm}p{2.0cm}p{1.4cm}p{1.4cm}p{1.4cm}p{1.4cm}p{1.4cm}}
    \toprule
        Benchmark & metric & Llama 2 70B & Gemini Pro 1.0 & Mistral Medium & Grok-1 & Reka Flash\\
        \midrule
        MMLU & $5$-shot & $68.9$ & $71.8$& $75.3$ & $73.0$ & $75.9$ \\ 
         GSM8K &maj@$8$ & $56.8$ & $86.5$ & $-$& $62.9$ & $85.8$ \\ 
        MATH & $4$-shot &  $13.5$& $32.6$ & $-$ & $23.9$ & $29.6$\\
        HumanEval & 0-shot &  $29.9$ & $67.7$ & $38.4$ & $63.2$ &  $72.0$ \\
        \midrule
        MMMU (vision) & 0-shot & N/A & $47.9$ & N/A & N/A & $53.3$\\
        VQAv2 & 0-shot & N/A & $77.2$ & N/A & N/A & $78.4$ \\ 
        Perception-test & 0-shot & N/A & $51.1$ & N/A & N/A & $56.4$ \\ 
       
        \bottomrule 
    \end{tabular}
 
\end{table}
Table \ref{tab:flash_results} reports results of Flash (21B) against other models of larger compute class. All competitors are approximately around 70B parameters with the exception of Grok-1 which is a sparse model with 314 billion parameters. We see that Flash outperforms (or is competitive to) all competitors on most benchmarks despite being much smaller.

\vspace{-0.5em}

\section{Conclusion}
\vspace{-0.5em}
We introduce a new series of powerful multimodal models, namely Reka \textbf{Core, Flash, Edge}. Reka \textbf{Flash} and \textbf{Edge} sets a new state-of-the-art on a compute-class basis, often delivering massive outsized value for their scale. Our \textbf{Core} model approaches frontier-class models on both human evaluation and automatic benchmarks. Reka \textbf{Core} is still improving so we expect to see even more improvements in the medium term. The field of large language models \citep{radford2018improving,brown2020language,devlin2018bert,raffel2019exploring,chowdhery2022palm,hoffmann2022training} is still nascent but moving very quickly. With that  comes the trade-off of significant amount of noise in the landscape. We hope this technical report shows the rigor of what it takes to build frontier-class models \textit{from scratch} given limited resources. 
\newpage

\newpage

\normalsize
\section{Appendix}

\subsection{MMMU breakdown}
In Table~\ref{tab:mmmu_category}, we report our category-level scores in MMMU~\citep{yue2023mmmu} for Reka Core.
\begin{table}[H]
    \caption{Breakdown of categories from MMMU benchmark \citep{yue2023mmmu}.}
    \centering
    \begin{tabular}{lc}
    \toprule
      Category   &  Score \\
      \midrule
         Art& $86.7$ \\ 
        Art Theory & $83.3$ \\ 
        Design & $86.7$ \\ 
        Music & $46.7$ \\ 
        Accounting & $46.7$\\ 
        Economics & $56.7$ \\
        Finance & $43.3$ \\ 
        Manage & $40.0$ \\ 
        Marketing & $50.0$\\
        Biology & $56.7$ \\
        Chemistry &$46.7$ \\
        Math & $46.7$ \\
        Physics &$36.7$ \\
        Basic Medical Science & $56.7$ \\
        Clinical Medicine & $60.0$ \\
        Diagnostics and Laboratory Medicine & $53.3$ \\
        Pharmacy & $63.3$ \\
        Public Health & $56.7$ \\
        History & $80.0$ \\
        Literature & $90.0$ \\ 
        Sociology & $73.3$ \\ 
        Agriculture & $70.0$ \\
        Architecture and Engineering & $40.0$ \\
        Computer Science & $50.0$ \\
        Electronics & $26.7$ \\
        Energy and Power & $43.3$ \\
        Materials & $36.7$ \\
        Mechanical Engineering & $43.3$ \\
        \midrule
        Overall & $56.3$ \\
        \bottomrule
      \end{tabular}
    \label{tab:mmmu_category}
\end{table}
\subsection{Historic versioning, changelog and timeline of Reka Chat models}
We include the version history of Reka models to easily refer to them across this tech report. 
\begin{table}[H]
    \centering
    \caption{Version history of all Reka Edge, Core and Flash models.}
    \begin{tabular}{llp{6cm}}
    \toprule
         Model&  Date & Comments\\
         \midrule 
         Reka Core v0.5 & Q2'24 & Apr launch version \\ 
         Reka Flash v1.5 & Q2'24 & Apr launch version \\
         Reka Flash v1.0 & Q1'24 & Feb public launch version \\
         Reka Edge v1.5 & Q2'24 & Apr launch version \\
         Reka Edge v1.0 & Q4'23 & Feb public launch version \\
         Reka Prototype v0.5 & Q3'23 & October private preview version \\
         \bottomrule
    \end{tabular}
    
    \label{tab:historic_versioning}
\end{table}

\newpage
\bibliographystyle{plainnat}
\newpage
\bibliography{ref}

\begin{thebibliography}{42}
\providecommand{\natexlab}[1]{#1}
\providecommand{\url}[1]{\texttt{#1}}
\expandafter\ifx\csname urlstyle\endcsname\relax
  \providecommand{\doi}[1]{doi: #1}\else
  \providecommand{\doi}{doi: \begingroup \urlstyle{rm}\Url}\fi

\bibitem[Ainslie et~al.(2023)Ainslie, Lee-Thorp, de~Jong, Zemlyanskiy, Lebrón, and Sanghai]{ainslie2023gqa}
Joshua Ainslie, James Lee-Thorp, Michiel de~Jong, Yury Zemlyanskiy, Federico Lebrón, and Sumit Sanghai.
\newblock Gqa: Training generalized multi-query transformer models from multi-head checkpoints, 2023.

\bibitem[Anthropic(2024)]{claude3}
Anthropic.
\newblock The claude 3 model family: Opus, sonnet, haiku.
\newblock 2024.

\bibitem[Artetxe et~al.(2019)Artetxe, Ruder, and Yogatama]{Artetxe:etal:2019}
Mikel Artetxe, Sebastian Ruder, and Dani Yogatama.
\newblock On the cross-lingual transferability of monolingual representations.
\newblock \emph{CoRR}, abs/1910.11856, 2019.

\bibitem[Askell et~al.(2021)Askell, Bai, Chen, Drain, Ganguli, Henighan, Jones, Joseph, Mann, DasSarma, Elhage, Hatfield-Dodds, Hernandez, Kernion, Ndousse, Olsson, Amodei, Brown, Clark, McCandlish, Olah, and Kaplan]{askell2021general}
Amanda Askell, Yuntao Bai, Anna Chen, Dawn Drain, Deep Ganguli, Tom Henighan, Andy Jones, Nicholas Joseph, Ben Mann, Nova DasSarma, Nelson Elhage, Zac Hatfield-Dodds, Danny Hernandez, Jackson Kernion, Kamal Ndousse, Catherine Olsson, Dario Amodei, Tom Brown, Jack Clark, Sam McCandlish, Chris Olah, and Jared Kaplan.
\newblock A general language assistant as a laboratory for alignment, 2021.

\bibitem[Bandarkar et~al.(2023)Bandarkar, Liang, Muller, Artetxe, Shukla, Husa, Goyal, Krishnan, Zettlemoyer, and Khabsa]{bandarkar2023belebele}
Lucas Bandarkar, Davis Liang, Benjamin Muller, Mikel Artetxe, Satya~Narayan Shukla, Donald Husa, Naman Goyal, Abhinandan Krishnan, Luke Zettlemoyer, and Madian Khabsa.
\newblock The belebele benchmark: a parallel reading comprehension dataset in 122 language variants, 2023.

\bibitem[Brown et~al.(2020)Brown, Mann, Ryder, Subbiah, Kaplan, Dhariwal, Neelakantan, Shyam, Sastry, Askell, et~al.]{brown2020language}
Tom~B Brown, Benjamin Mann, Nick Ryder, Melanie Subbiah, Jared Kaplan, Prafulla Dhariwal, Arvind Neelakantan, Pranav Shyam, Girish Sastry, Amanda Askell, et~al.
\newblock Language models are few-shot learners.
\newblock \emph{arXiv preprint arXiv:2005.14165}, 2020.

\bibitem[Chen et~al.(2021)Chen, Tworek, Jun, Yuan, de~Oliveira~Pinto, Kaplan, Edwards, Burda, Joseph, Brockman, Ray, Puri, Krueger, Petrov, Khlaaf, Sastry, Mishkin, Chan, Gray, Ryder, Pavlov, Power, Kaiser, Bavarian, Winter, Tillet, Such, Cummings, Plappert, Chantzis, Barnes, Herbert-Voss, Guss, Nichol, Paino, Tezak, Tang, Babuschkin, Balaji, Jain, Saunders, Hesse, Carr, Leike, Achiam, Misra, Morikawa, Radford, Knight, Brundage, Murati, Mayer, Welinder, McGrew, Amodei, McCandlish, Sutskever, and Zaremba]{chen2021codex}
Mark Chen, Jerry Tworek, Heewoo Jun, Qiming Yuan, Henrique~Ponde de~Oliveira~Pinto, Jared Kaplan, Harri Edwards, Yuri Burda, Nicholas Joseph, Greg Brockman, Alex Ray, Raul Puri, Gretchen Krueger, Michael Petrov, Heidy Khlaaf, Girish Sastry, Pamela Mishkin, Brooke Chan, Scott Gray, Nick Ryder, Mikhail Pavlov, Alethea Power, Lukasz Kaiser, Mohammad Bavarian, Clemens Winter, Philippe Tillet, Felipe~Petroski Such, Dave Cummings, Matthias Plappert, Fotios Chantzis, Elizabeth Barnes, Ariel Herbert-Voss, William~Hebgen Guss, Alex Nichol, Alex Paino, Nikolas Tezak, Jie Tang, Igor Babuschkin, Suchir Balaji, Shantanu Jain, William Saunders, Christopher Hesse, Andrew~N. Carr, Jan Leike, Josh Achiam, Vedant Misra, Evan Morikawa, Alec Radford, Matthew Knight, Miles Brundage, Mira Murati, Katie Mayer, Peter Welinder, Bob McGrew, Dario Amodei, Sam McCandlish, Ilya Sutskever, and Wojciech Zaremba.
\newblock Evaluating large language models trained on code.
\newblock 2021.

\bibitem[Chen et~al.(2023)Chen, Cano, Romanou, Bonnet, Matoba, Salvi, Pagliardini, Fan, Köpf, Mohtashami, Sallinen, Sakhaeirad, Swamy, Krawczuk, Bayazit, Marmet, Montariol, Hartley, Jaggi, and Bosselut]{chen2023meditron70b}
Zeming Chen, Alejandro~Hernández Cano, Angelika Romanou, Antoine Bonnet, Kyle Matoba, Francesco Salvi, Matteo Pagliardini, Simin Fan, Andreas Köpf, Amirkeivan Mohtashami, Alexandre Sallinen, Alireza Sakhaeirad, Vinitra Swamy, Igor Krawczuk, Deniz Bayazit, Axel Marmet, Syrielle Montariol, Mary-Anne Hartley, Martin Jaggi, and Antoine Bosselut.
\newblock Meditron-70b: Scaling medical pretraining for large language models, 2023.

\bibitem[Chowdhery et~al.(2022)Chowdhery, Narang, Devlin, Bosma, Mishra, Roberts, Barham, Chung, Sutton, Gehrmann, et~al.]{chowdhery2022palm}
Aakanksha Chowdhery, Sharan Narang, Jacob Devlin, Maarten Bosma, Gaurav Mishra, Adam Roberts, Paul Barham, Hyung~Won Chung, Charles Sutton, Sebastian Gehrmann, et~al.
\newblock Palm: Scaling language modeling with pathways.
\newblock \emph{arXiv preprint arXiv:2204.02311}, 2022.

\bibitem[Chung et~al.(2024)Chung, Hou, Longpre, Zoph, Tay, Fedus, Li, Wang, Dehghani, Brahma, et~al.]{chung2022flan}
Hyung~Won Chung, Le~Hou, Shayne Longpre, Barret Zoph, Yi~Tay, William Fedus, Yunxuan Li, Xuezhi Wang, Mostafa Dehghani, Siddhartha Brahma, et~al.
\newblock Scaling instruction-finetuned language models.
\newblock \emph{Journal of Machine Learning Research}, 25\penalty0 (70):\penalty0 1--53, 2024.

\bibitem[Clark et~al.(2020)Clark, Choi, Collins, Garrette, Kwiatkowski, Nikolaev, and Palomaki]{clark2020tydi}
Jonathan~H Clark, Eunsol Choi, Michael Collins, Dan Garrette, Tom Kwiatkowski, Vitaly Nikolaev, and Jennimaria Palomaki.
\newblock Tydi qa: A benchmark for information-seeking question answering in typologically diverse languages.
\newblock \emph{Transactions of the Association for Computational Linguistics}, 8:\penalty0 454--470, 2020.

\bibitem[Cobbe et~al.(2021)Cobbe, Kosaraju, Bavarian, Chen, Jun, Kaiser, Plappert, Tworek, Hilton, Nakano, Hesse, and Schulman]{cobbe2021training}
Karl Cobbe, Vineet Kosaraju, Mohammad Bavarian, Mark Chen, Heewoo Jun, Lukasz Kaiser, Matthias Plappert, Jerry Tworek, Jacob Hilton, Reiichiro Nakano, Christopher Hesse, and John Schulman.
\newblock Training verifiers to solve math word problems, 2021.

\bibitem[Devlin et~al.(2018)Devlin, Chang, Lee, and Toutanova]{devlin2018bert}
Jacob Devlin, Ming-Wei Chang, Kenton Lee, and Kristina Toutanova.
\newblock Bert: Pre-training of deep bidirectional transformers for language understanding.
\newblock \emph{arXiv preprint arXiv:1810.04805}, 2018.

\bibitem[Gemma et~al.(2024)Gemma, Mesnard, Hardin, Dadashi, Bhupatiraju, Pathak, Sifre, Rivière, Kale, Love, Tafti, Hussenot, Sessa, Chowdhery, Roberts, Barua, Botev, Castro-Ros, Slone, Héliou, Tacchetti, Bulanova, Paterson, Tsai, Shahriari, Lan, Choquette-Choo, Crepy, Cer, Ippolito, Reid, Buchatskaya, Ni, Noland, Yan, Tucker, Muraru, Rozhdestvenskiy, Michalewski, Tenney, Grishchenko, Austin, Keeling, Labanowski, Lespiau, Stanway, Brennan, Chen, Ferret, Chiu, Mao-Jones, Lee, Yu, Millican, Sjoesund, Lee, Dixon, Reid, Mikuła, Wirth, Sharman, Chinaev, Thain, Bachem, Chang, Wahltinez, Bailey, Michel, Yotov, Chaabouni, Comanescu, Jana, Anil, McIlroy, Liu, Mullins, Smith, Borgeaud, Girgin, Douglas, Pandya, Shakeri, De, Klimenko, Hennigan, Feinberg, Stokowiec, hui Chen, Ahmed, Gong, Warkentin, Peran, Giang, Farabet, Vinyals, Dean, Kavukcuoglu, Hassabis, Ghahramani, Eck, Barral, Pereira, Collins, Joulin, Fiedel, Senter, Andreev, and Kenealy]{gemmateam2024gemma}
Gemma, Thomas Mesnard, Cassidy Hardin, Robert Dadashi, Surya Bhupatiraju, Shreya Pathak, Laurent Sifre, Morgane Rivière, Mihir~Sanjay Kale, Juliette Love, Pouya Tafti, Léonard Hussenot, Pier~Giuseppe Sessa, Aakanksha Chowdhery, Adam Roberts, Aditya Barua, Alex Botev, Alex Castro-Ros, Ambrose Slone, Amélie Héliou, Andrea Tacchetti, Anna Bulanova, Antonia Paterson, Beth Tsai, Bobak Shahriari, Charline~Le Lan, Christopher~A. Choquette-Choo, Clément Crepy, Daniel Cer, Daphne Ippolito, David Reid, Elena Buchatskaya, Eric Ni, Eric Noland, Geng Yan, George Tucker, George-Christian Muraru, Grigory Rozhdestvenskiy, Henryk Michalewski, Ian Tenney, Ivan Grishchenko, Jacob Austin, James Keeling, Jane Labanowski, Jean-Baptiste Lespiau, Jeff Stanway, Jenny Brennan, Jeremy Chen, Johan Ferret, Justin Chiu, Justin Mao-Jones, Katherine Lee, Kathy Yu, Katie Millican, Lars~Lowe Sjoesund, Lisa Lee, Lucas Dixon, Machel Reid, Maciej Mikuła, Mateo Wirth, Michael Sharman, Nikolai Chinaev, Nithum Thain, Olivier Bachem, Oscar
  Chang, Oscar Wahltinez, Paige Bailey, Paul Michel, Petko Yotov, Rahma Chaabouni, Ramona Comanescu, Reena Jana, Rohan Anil, Ross McIlroy, Ruibo Liu, Ryan Mullins, Samuel~L Smith, Sebastian Borgeaud, Sertan Girgin, Sholto Douglas, Shree Pandya, Siamak Shakeri, Soham De, Ted Klimenko, Tom Hennigan, Vlad Feinberg, Wojciech Stokowiec, Yu~hui Chen, Zafarali Ahmed, Zhitao Gong, Tris Warkentin, Ludovic Peran, Minh Giang, Clément Farabet, Oriol Vinyals, Jeff Dean, Koray Kavukcuoglu, Demis Hassabis, Zoubin Ghahramani, Douglas Eck, Joelle Barral, Fernando Pereira, Eli Collins, Armand Joulin, Noah Fiedel, Evan Senter, Alek Andreev, and Kathleen Kenealy.
\newblock Gemma: Open models based on gemini research and technology, 2024.

\bibitem[Google et~al.(2023)Google, Anil, Dai, Firat, Johnson, Lepikhin, Passos, Shakeri, Taropa, Bailey, Chen, Chu, Clark, Shafey, Huang, Meier-Hellstern, Mishra, Moreira, Omernick, Robinson, Ruder, Tay, Xiao, Xu, Zhang, Abrego, Ahn, Austin, Barham, Botha, Bradbury, Brahma, Brooks, Catasta, Cheng, Cherry, Choquette-Choo, Chowdhery, Crepy, Dave, Dehghani, Dev, Devlin, Díaz, Du, Dyer, Feinberg, Feng, Fienber, Freitag, Garcia, Gehrmann, Gonzalez, Gur-Ari, Hand, Hashemi, Hou, Howland, Hu, Hui, Hurwitz, Isard, Ittycheriah, Jagielski, Jia, Kenealy, Krikun, Kudugunta, Lan, Lee, Lee, Li, Li, Li, Li, Li, Lim, Lin, Liu, Liu, Maggioni, Mahendru, Maynez, Misra, Moussalem, Nado, Nham, Ni, Nystrom, Parrish, Pellat, Polacek, Polozov, Pope, Qiao, Reif, Richter, Riley, Ros, Roy, Saeta, Samuel, Shelby, Slone, Smilkov, So, Sohn, Tokumine, Valter, Vasudevan, Vodrahalli, Wang, Wang, Wang, Wang, Wieting, Wu, Xu, Xu, Xue, Yin, Yu, Zhang, Zheng, Zheng, Zhou, Zhou, Petrov, and Wu]{google2023palm}
Google, Rohan Anil, Andrew~M. Dai, Orhan Firat, Melvin Johnson, Dmitry Lepikhin, Alexandre Passos, Siamak Shakeri, Emanuel Taropa, Paige Bailey, Zhifeng Chen, Eric Chu, Jonathan~H. Clark, Laurent~El Shafey, Yanping Huang, Kathy Meier-Hellstern, Gaurav Mishra, Erica Moreira, Mark Omernick, Kevin Robinson, Sebastian Ruder, Yi~Tay, Kefan Xiao, Yuanzhong Xu, Yujing Zhang, Gustavo~Hernandez Abrego, Junwhan Ahn, Jacob Austin, Paul Barham, Jan Botha, James Bradbury, Siddhartha Brahma, Kevin Brooks, Michele Catasta, Yong Cheng, Colin Cherry, Christopher~A. Choquette-Choo, Aakanksha Chowdhery, Clément Crepy, Shachi Dave, Mostafa Dehghani, Sunipa Dev, Jacob Devlin, Mark Díaz, Nan Du, Ethan Dyer, Vlad Feinberg, Fangxiaoyu Feng, Vlad Fienber, Markus Freitag, Xavier Garcia, Sebastian Gehrmann, Lucas Gonzalez, Guy Gur-Ari, Steven Hand, Hadi Hashemi, Le~Hou, Joshua Howland, Andrea Hu, Jeffrey Hui, Jeremy Hurwitz, Michael Isard, Abe Ittycheriah, Matthew Jagielski, Wenhao Jia, Kathleen Kenealy, Maxim Krikun, Sneha
  Kudugunta, Chang Lan, Katherine Lee, Benjamin Lee, Eric Li, Music Li, Wei Li, YaGuang Li, Jian Li, Hyeontaek Lim, Hanzhao Lin, Zhongtao Liu, Frederick Liu, Marcello Maggioni, Aroma Mahendru, Joshua Maynez, Vedant Misra, Maysam Moussalem, Zachary Nado, John Nham, Eric Ni, Andrew Nystrom, Alicia Parrish, Marie Pellat, Martin Polacek, Alex Polozov, Reiner Pope, Siyuan Qiao, Emily Reif, Bryan Richter, Parker Riley, Alex~Castro Ros, Aurko Roy, Brennan Saeta, Rajkumar Samuel, Renee Shelby, Ambrose Slone, Daniel Smilkov, David~R. So, Daniel Sohn, Simon Tokumine, Dasha Valter, Vijay Vasudevan, Kiran Vodrahalli, Xuezhi Wang, Pidong Wang, Zirui Wang, Tao Wang, John Wieting, Yuhuai Wu, Kelvin Xu, Yunhan Xu, Linting Xue, Pengcheng Yin, Jiahui Yu, Qiao Zhang, Steven Zheng, Ce~Zheng, Weikang Zhou, Denny Zhou, Slav Petrov, and Yonghui Wu.
\newblock Palm 2 technical report, 2023.

\bibitem[Google(2023)]{geminiteam2023gemini_short}
Gemini~Team Google.
\newblock Gemini: A family of highly capable multimodal models, 2023.

\bibitem[Goyal et~al.(2017)Goyal, Khot, Summers{-}Stay, Batra, and Parikh]{balanced_vqa_v2}
Yash Goyal, Tejas Khot, Douglas Summers{-}Stay, Dhruv Batra, and Devi Parikh.
\newblock Making the {V} in {VQA} matter: Elevating the role of image understanding in {V}isual {Q}uestion {A}nswering.
\newblock In \emph{Conference on Computer Vision and Pattern Recognition (CVPR)}, 2017.

\bibitem[Hendrycks et~al.(2021)Hendrycks, Burns, Basart, Zou, Mazeika, Song, and Steinhardt]{hendrycks2021measuring}
Dan Hendrycks, Collin Burns, Steven Basart, Andy Zou, Mantas Mazeika, Dawn Song, and Jacob Steinhardt.
\newblock Measuring massive multitask language understanding, 2021.

\bibitem[Hoffmann et~al.(2022)Hoffmann, Borgeaud, Mensch, Buchatskaya, Cai, Rutherford, Casas, Hendricks, Welbl, Clark, et~al.]{hoffmann2022training}
Jordan Hoffmann, Sebastian Borgeaud, Arthur Mensch, Elena Buchatskaya, Trevor Cai, Eliza Rutherford, Diego de~Las Casas, Lisa~Anne Hendricks, Johannes Welbl, Aidan Clark, et~al.
\newblock Training compute-optimal large language models.
\newblock \emph{arXiv preprint arXiv:2203.15556}, 2022.

\bibitem[Jiang et~al.(2023)Jiang, Sablayrolles, Mensch, Bamford, Chaplot, de~las Casas, Bressand, Lengyel, Lample, Saulnier, Lavaud, Lachaux, Stock, Scao, Lavril, Wang, Lacroix, and Sayed]{jiang2023mistral}
Albert~Q. Jiang, Alexandre Sablayrolles, Arthur Mensch, Chris Bamford, Devendra~Singh Chaplot, Diego de~las Casas, Florian Bressand, Gianna Lengyel, Guillaume Lample, Lucile Saulnier, Lélio~Renard Lavaud, Marie-Anne Lachaux, Pierre Stock, Teven~Le Scao, Thibaut Lavril, Thomas Wang, Timothée Lacroix, and William~El Sayed.
\newblock Mistral 7b, 2023.

\bibitem[Lin et~al.(2022)Lin, Mihaylov, Artetxe, Wang, Chen, Simig, Ott, Goyal, Bhosale, Du, Pasunuru, Shleifer, Koura, Chaudhary, O{'}Horo, Wang, Zettlemoyer, Kozareva, Diab, Stoyanov, and Li]{lin-etal-2022-shot}
Xi~Victoria Lin, Todor Mihaylov, Mikel Artetxe, Tianlu Wang, Shuohui Chen, Daniel Simig, Myle Ott, Naman Goyal, Shruti Bhosale, Jingfei Du, Ramakanth Pasunuru, Sam Shleifer, Punit~Singh Koura, Vishrav Chaudhary, Brian O{'}Horo, Jeff Wang, Luke Zettlemoyer, Zornitsa Kozareva, Mona Diab, Veselin Stoyanov, and Xian Li.
\newblock Few-shot learning with multilingual generative language models.
\newblock In Yoav Goldberg, Zornitsa Kozareva, and Yue Zhang, editors, \emph{Proceedings of the 2022 Conference on Empirical Methods in Natural Language Processing}, pages 9019--9052, Abu Dhabi, United Arab Emirates, December 2022. Association for Computational Linguistics.
\newblock \doi{10.18653/v1/2022.emnlp-main.616}.
\newblock URL \url{https://aclanthology.org/2022.emnlp-main.616}.

\bibitem[OpenAI(2023)]{openai2023gpt4}
OpenAI.
\newblock Gpt-4 technical report, 2023.

\bibitem[OpenAI(2024)]{gpt4v}
OpenAI.
\newblock Gpt-4v(ision) system card.
\newblock 2024.

\bibitem[Ouyang et~al.(2022)Ouyang, Wu, Jiang, Almeida, Wainwright, Mishkin, Zhang, Agarwal, Slama, Ray, Schulman, Hilton, Kelton, Miller, Simens, Askell, Welinder, Christiano, Leike, and Lowe]{ouyang2022training}
Long Ouyang, Jeff Wu, Xu~Jiang, Diogo Almeida, Carroll~L. Wainwright, Pamela Mishkin, Chong Zhang, Sandhini Agarwal, Katarina Slama, Alex Ray, John Schulman, Jacob Hilton, Fraser Kelton, Luke Miller, Maddie Simens, Amanda Askell, Peter Welinder, Paul Christiano, Jan Leike, and Ryan Lowe.
\newblock Training language models to follow instructions with human feedback, 2022.

\bibitem[Paszke et~al.(2019)Paszke, Gross, Massa, Lerer, Bradbury, Chanan, Killeen, Lin, Gimelshein, Antiga, Desmaison, K{\"{o}}pf, Yang, DeVito, Raison, Tejani, Chilamkurthy, Steiner, Fang, Bai, and Chintala]{pytorch}
Adam Paszke, Sam Gross, Francisco Massa, Adam Lerer, James Bradbury, Gregory Chanan, Trevor Killeen, Zeming Lin, Natalia Gimelshein, Luca Antiga, Alban Desmaison, Andreas K{\"{o}}pf, Edward~Z. Yang, Zach DeVito, Martin Raison, Alykhan Tejani, Sasank Chilamkurthy, Benoit Steiner, Lu~Fang, Junjie Bai, and Soumith Chintala.
\newblock Pytorch: An imperative style, high-performance deep learning library.
\newblock \emph{CoRR}, abs/1912.01703, 2019.
\newblock URL \url{http://arxiv.org/abs/1912.01703}.

\bibitem[Ponti et~al.(2020)Ponti, Glava{\v{s}}, Majewska, Liu, Vuli{\'c}, and Korhonen]{ponti-etal-2020-xcopa}
Edoardo~Maria Ponti, Goran Glava{\v{s}}, Olga Majewska, Qianchu Liu, Ivan Vuli{\'c}, and Anna Korhonen.
\newblock {XCOPA}: A multilingual dataset for causal commonsense reasoning.
\newblock In Bonnie Webber, Trevor Cohn, Yulan He, and Yang Liu, editors, \emph{Proceedings of the 2020 Conference on Empirical Methods in Natural Language Processing (EMNLP)}, pages 2362--2376, Online, November 2020. Association for Computational Linguistics.
\newblock \doi{10.18653/v1/2020.emnlp-main.185}.
\newblock URL \url{https://aclanthology.org/2020.emnlp-main.185}.

\bibitem[Pătrăucean et~al.(2023)Pătrăucean, Smaira, Gupta, Continente, Markeeva, Banarse, Koppula, Heyward, Malinowski, Yang, Doersch, Matejovicova, Sulsky, Miech, Frechette, Klimczak, Koster, Zhang, Winkler, Aytar, Osindero, Damen, Zisserman, and Carreira]{pătrăucean2023perception}
Viorica Pătrăucean, Lucas Smaira, Ankush Gupta, Adrià~Recasens Continente, Larisa Markeeva, Dylan Banarse, Skanda Koppula, Joseph Heyward, Mateusz Malinowski, Yi~Yang, Carl Doersch, Tatiana Matejovicova, Yury Sulsky, Antoine Miech, Alex Frechette, Hanna Klimczak, Raphael Koster, Junlin Zhang, Stephanie Winkler, Yusuf Aytar, Simon Osindero, Dima Damen, Andrew Zisserman, and João Carreira.
\newblock Perception test: A diagnostic benchmark for multimodal video models, 2023.

\bibitem[Radford et~al.(2018)Radford, Narasimhan, Salimans, Sutskever, et~al.]{radford2018improving}
Alec Radford, Karthik Narasimhan, Tim Salimans, Ilya Sutskever, et~al.
\newblock Improving language understanding by generative pre-training.
\newblock 2018.

\bibitem[Raffel et~al.(2019)Raffel, Shazeer, Roberts, Lee, Narang, Matena, Zhou, Li, and Liu]{raffel2019exploring}
Colin Raffel, Noam Shazeer, Adam Roberts, Katherine Lee, Sharan Narang, Michael Matena, Yanqi Zhou, Wei Li, and Peter~J Liu.
\newblock Exploring the limits of transfer learning with a unified text-to-text transformer.
\newblock \emph{arXiv preprint arXiv:1910.10683}, 2019.

\bibitem[Rein et~al.(2023)Rein, Hou, Stickland, Petty, Pang, Dirani, Michael, and Bowman]{rein2023gpqa}
David Rein, Betty~Li Hou, Asa~Cooper Stickland, Jackson Petty, Richard~Yuanzhe Pang, Julien Dirani, Julian Michael, and Samuel~R. Bowman.
\newblock Gpqa: A graduate-level google-proof q\&a benchmark, 2023.

\bibitem[Schulman et~al.(2017)Schulman, Wolski, Dhariwal, Radford, and Klimov]{ppo}
John Schulman, Filip Wolski, Prafulla Dhariwal, Alec Radford, and Oleg Klimov.
\newblock Proximal policy optimization algorithms.
\newblock \emph{CoRR}, abs/1707.06347, 2017.
\newblock URL \url{http://arxiv.org/abs/1707.06347}.

\bibitem[Shazeer(2019)]{shazeer2019fast}
Noam Shazeer.
\newblock Fast transformer decoding: One write-head is all you need.
\newblock \emph{arXiv preprint arXiv:1911.02150}, 2019.

\bibitem[Shazeer(2020)]{shazeer2020glu}
Noam Shazeer.
\newblock Glu variants improve transformer.
\newblock \emph{arXiv preprint arXiv:2002.05202}, 2020.

\bibitem[Singhal et~al.(2023)Singhal, Tu, Gottweis, Sayres, Wulczyn, Hou, Clark, Pfohl, Cole-Lewis, Neal, Schaekermann, Wang, Amin, Lachgar, Mansfield, Prakash, Green, Dominowska, y~Arcas, Tomasev, Liu, Wong, Semturs, Mahdavi, Barral, Webster, Corrado, Matias, Azizi, Karthikesalingam, and Natarajan]{singhal2023expertlevel}
Karan Singhal, Tao Tu, Juraj Gottweis, Rory Sayres, Ellery Wulczyn, Le~Hou, Kevin Clark, Stephen Pfohl, Heather Cole-Lewis, Darlene Neal, Mike Schaekermann, Amy Wang, Mohamed Amin, Sami Lachgar, Philip Mansfield, Sushant Prakash, Bradley Green, Ewa Dominowska, Blaise~Aguera y~Arcas, Nenad Tomasev, Yun Liu, Renee Wong, Christopher Semturs, S.~Sara Mahdavi, Joelle Barral, Dale Webster, Greg~S. Corrado, Yossi Matias, Shekoofeh Azizi, Alan Karthikesalingam, and Vivek Natarajan.
\newblock Towards expert-level medical question answering with large language models, 2023.

\bibitem[Su et~al.(2021)Su, Lu, Pan, Wen, and Liu]{rope-paper}
Jianlin Su, Yu~Lu, Shengfeng Pan, Bo~Wen, and Yunfeng Liu.
\newblock Roformer: Enhanced transformer with rotary position embedding.
\newblock \emph{arXiv preprint arXiv:2104.09864}, 2021.

\bibitem[Tay(2024)]{tayblog}
Yi~Tay.
\newblock Training great llms entirely from ground up in the wilderness as a startup.
\newblock 2024.

\bibitem[Touvron et~al.(2023)Touvron, Martin, Stone, Albert, Almahairi, Babaei, Bashlykov, Batra, Bhargava, Bhosale, Bikel, Blecher, Ferrer, Chen, Cucurull, Esiobu, Fernandes, Fu, Fu, Fuller, Gao, Goswami, Goyal, Hartshorn, Hosseini, Hou, Inan, Kardas, Kerkez, Khabsa, Kloumann, Korenev, Koura, Lachaux, Lavril, Lee, Liskovich, Lu, Mao, Martinet, Mihaylov, Mishra, Molybog, Nie, Poulton, Reizenstein, Rungta, Saladi, Schelten, Silva, Smith, Subramanian, Tan, Tang, Taylor, Williams, Kuan, Xu, Yan, Zarov, Zhang, Fan, Kambadur, Narang, Rodriguez, Stojnic, Edunov, and Scialom]{touvron2023llama}
Hugo Touvron, Louis Martin, Kevin Stone, Peter Albert, Amjad Almahairi, Yasmine Babaei, Nikolay Bashlykov, Soumya Batra, Prajjwal Bhargava, Shruti Bhosale, Dan Bikel, Lukas Blecher, Cristian~Canton Ferrer, Moya Chen, Guillem Cucurull, David Esiobu, Jude Fernandes, Jeremy Fu, Wenyin Fu, Brian Fuller, Cynthia Gao, Vedanuj Goswami, Naman Goyal, Anthony Hartshorn, Saghar Hosseini, Rui Hou, Hakan Inan, Marcin Kardas, Viktor Kerkez, Madian Khabsa, Isabel Kloumann, Artem Korenev, Punit~Singh Koura, Marie-Anne Lachaux, Thibaut Lavril, Jenya Lee, Diana Liskovich, Yinghai Lu, Yuning Mao, Xavier Martinet, Todor Mihaylov, Pushkar Mishra, Igor Molybog, Yixin Nie, Andrew Poulton, Jeremy Reizenstein, Rashi Rungta, Kalyan Saladi, Alan Schelten, Ruan Silva, Eric~Michael Smith, Ranjan Subramanian, Xiaoqing~Ellen Tan, Binh Tang, Ross Taylor, Adina Williams, Jian~Xiang Kuan, Puxin Xu, Zheng Yan, Iliyan Zarov, Yuchen Zhang, Angela Fan, Melanie Kambadur, Sharan Narang, Aurelien Rodriguez, Robert Stojnic, Sergey Edunov, and Thomas
  Scialom.
\newblock Llama 2: Open foundation and fine-tuned chat models, 2023.

\bibitem[Wei et~al.(2021)Wei, Bosma, Zhao, Guu, Yu, Lester, Du, Dai, and Le]{wei2021finetuned}
Jason Wei, Maarten Bosma, Vincent~Y Zhao, Kelvin Guu, Adams~Wei Yu, Brian Lester, Nan Du, Andrew~M Dai, and Quoc~V Le.
\newblock Finetuned language models are zero-shot learners.
\newblock \emph{arXiv preprint arXiv:2109.01652}, 2021.

\bibitem[Wei et~al.(2022)Wei, Wang, Schuurmans, Bosma, Ichter, Xia, Chi, Le, and Zhou]{wei2022chain}
Jason Wei, Xuezhi Wang, Dale Schuurmans, Maarten Bosma, Brian Ichter, Fei Xia, Ed~Chi, Quoc Le, and Denny Zhou.
\newblock Chain of thought prompting elicits reasoning in large language models.
\newblock \emph{Conference on Neural Information Processing Systems (NeurIPS)}, 2022.

\bibitem[xAI(2023)]{grok}
xAI.
\newblock Announcing grok.
\newblock 2023.

\bibitem[Yue et~al.(2024)Yue, Ni, Zhang, Zheng, Liu, Zhang, Stevens, Jiang, Ren, Sun, Wei, Yu, Yuan, Sun, Yin, Zheng, Yang, Liu, Huang, Sun, Su, and Chen]{yue2023mmmu}
Xiang Yue, Yuansheng Ni, Kai Zhang, Tianyu Zheng, Ruoqi Liu, Ge~Zhang, Samuel Stevens, Dongfu Jiang, Weiming Ren, Yuxuan Sun, Cong Wei, Botao Yu, Ruibin Yuan, Renliang Sun, Ming Yin, Boyuan Zheng, Zhenzhu Yang, Yibo Liu, Wenhao Huang, Huan Sun, Yu~Su, and Wenhu Chen.
\newblock Mmmu: A massive multi-discipline multimodal understanding and reasoning benchmark for expert agi.
\newblock In \emph{Proceedings of CVPR}, 2024.

\bibitem[Zhang and Sennrich(2019)]{zhang2019root}
Biao Zhang and Rico Sennrich.
\newblock Root mean square layer normalization, 2019.

\end{thebibliography}
\newpage

\end{document}